\documentclass[10pt, conference]{IEEEtran}

\usepackage{amsmath,amssymb,amsfonts}
\usepackage{graphicx}
\usepackage{pifont}
\usepackage{tikz}

\usepackage{amsthm}
\usepackage{amsmath}
\usepackage{xcolor}
\usepackage[figuresright]{rotating}

\usepackage{graphicx}
\graphicspath{{/}{fig/}}

\usepackage{array}
\usepackage{textcomp}
\usepackage{xcolor}
\usepackage{multirow}
\usepackage{booktabs}

\usepackage{mathtools}
\usepackage{breqn}
\usepackage{float}

\usepackage{fancyhdr}

\usepackage{pgfplots}
\pgfplotsset{compat=1.7}
\usepgfplotslibrary{groupplots}

\usepackage{silence}
\usepackage{tabularx}
\usepackage{hyperref}
\usepackage{flushend}
\usepackage{algorithmic}
\usepackage[linesnumbered,ruled,vlined,shortend]{algorithm2e}
\usepackage{subcaption}

\newlength\figureheight
\newlength\figurewidth
\SetAlCapNameFnt{\footnotesize}
\SetAlCapFnt{\footnotesize}

\definecolor{bestcolor}{RGB}{255, 69, 0} 
\definecolor{secondcolor}{RGB}{255, 165, 0} 

\title{
    Dual-Criterion Model Aggregation in Federated Learning: Balancing Data Quantity and Quality \\
}

\author{
    \IEEEauthorblockN{
        \vspace{1em}
        Haizhou Zhang\IEEEauthorrefmark{2},
        Xianjia Yu\IEEEauthorrefmark{2},
        Tomi Westerlund\IEEEauthorrefmark{2}
    }
    \IEEEauthorblockA{
        \normalsize
        \IEEEauthorrefmark{2}\href{https://tiers.utu.fi}{Turku Intelligent Embedded and Robotic Systems (TIERS) Lab, University of Turku, Finland}.\\
        Emails: \textsuperscript{1}\{hazhan, xianjia.yu, tovewe\}@utu.fi\\[+6pt]
    }
}

\fancypagestyle{FirstPage}{

}

\begin{document}

\maketitle
\thispagestyle{empty}
\pagestyle{empty}



\begin{abstract}%
    \label{sec:abstract}
    Federated learning (FL) has become one of the key methods for privacy-preserving collaborative learning, as it enables the transfer of models without requiring local data exchange. Within the FL framework, an aggregation algorithm is recognized as one of the most crucial components for ensuring the efficacy and security of the system.
    Existing average aggregation algorithms typically assume that all client-trained data holds equal value or that weights are based solely on the quantity of data contributed by each client. In contrast, alternative approaches involve training the model locally after aggregation to enhance adaptability. However, these approaches fundamentally ignore the inherent heterogeneity between different clients' data and the complexity of variations in data at the aggregation stage, which may lead to a suboptimal global model. 
    
    To address these issues, this study proposes a novel dual-criterion weighted aggregation algorithm involving the quantity and quality of data from the client node. Specifically, we quantify the data used for training and perform multiple rounds of local model inference accuracy evaluation on a specialized dataset to assess the data quality of each client. 
    These two factors are utilized as weights within the aggregation process, applied through a dynamically weighted summation of these two factors.
    This approach allows the algorithm to adaptively adjust the weights, ensuring that every client can contribute to the global model, regardless of their data's size or initial quality. Our experiments show that the proposed algorithm outperforms several existing state-of-the-art aggregation approaches on both a general-purpose open-source dataset, CIFAR-10, and a dataset specific to visual obstacle avoidance.

\end{abstract}

\IEEEpeerreviewmaketitle

\section{Introduction}
\label{sec:introduction}

In the last decade, deep learning (DL) has revolutionized a wide array of sectors. However, these AI algorithms are often dependent on massive and diverse datasets for training, which imposes a computational burden and raises significant privacy concerns, especially for end-users who are reluctant to share their raw data with centralized servers. 
For example, in the field of robotics, where achieving high-level autonomy necessitates the training of DL models on the sensor data transmitted across networked robots to enhance collaborative situational awareness. However, in certain cases, the location or context of the robots raises privacy concerns that restrict raw data sharing.

Federated learning (FL) has emerged as a revolutionary approach, enabling collaborative model training across different decentralized nodes without compromising data privacy or security~\cite{XIANJIA2021135}~\cite{yu2022federated}. FL allows decentralized collaborative learning by aggregating the local model updates from clients into a coherent global model without data transmission. The aggregation algorithm is the key component of FL.

However, existing aggregation methods, such as Federated Averaging (FedAvg)~\cite{McMahan2016CommunicationEfficientLO}, predominantly weigh models based on the quantity of data contributed by each client, overlooking the heterogeneity in data quality and distribution across nodes. This oversight can lead to suboptimal global models, particularly in scenarios where data relevance and quality significantly vary. This is more obvious when the data among clients are imbalanced or in non-IID (independently and identically distributed).

\begin{figure}
    \centering
    \includegraphics[width=0.49\textwidth]{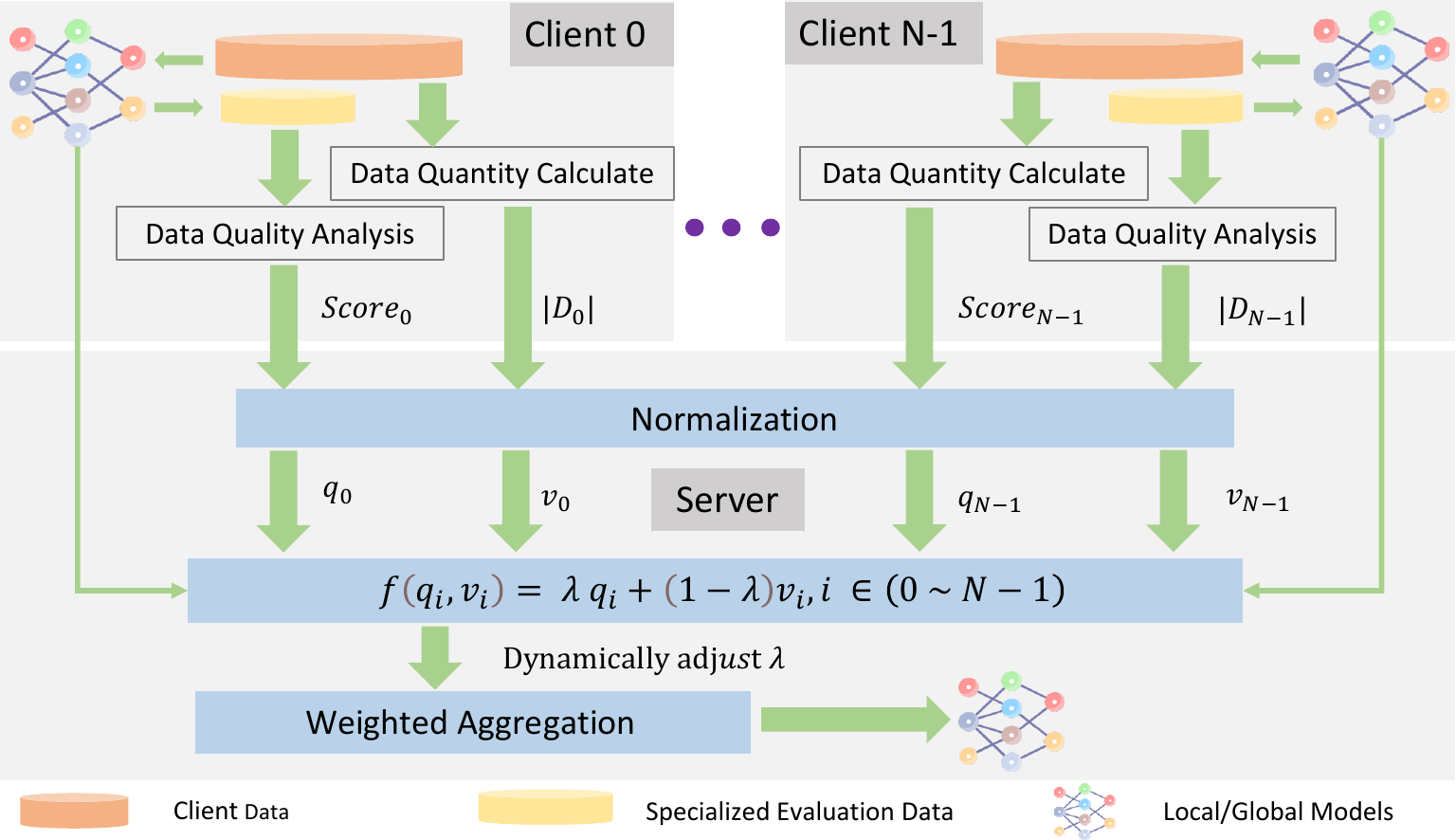}
    \caption{Conceptual diagram of our proposed dual criterion model aggregation approach}
    \label{fig:dual-criterion}
\end{figure}

To address the aforementioned issues, this paper introduces a novel dual-criterion aggregation mechanism as shown in Fig.~\ref{fig:dual-criterion}. In the aggregation process, we calculate the quantity of the training dataset and conduct local model inference accuracy evaluation on a specialized dataset to assess the data quality of each client on a separate evaluation dataset. These two factors are then used as weights in the aggregation. This approach also enables the algorithm to dynamically adjust weights by a ratio ensuring that each client contributes proportionally to the global model, irrespective of the size or initial quality of their data.

We validated our experimental results on two datasets: the dataset for visual obstacle avoidance from our previous work~\cite{yu2022federated} and the CIFAR-10 dataset~\cite{krizhevsky2009learning}. 
The results demonstrate that the proposed method performs exceptionally well across various scenarios proving that our dual-criterion method can optimize the learning process. This enhancement of the global model's performance is achieved by prioritizing high-quality, relevant data in the aggregation process.
Furthermore, to support broader scholarly investigation, we have also open-sourced our work on \href{https://github.com/Jimmy-Zhang13798/A-new-weighted-aggregation-method}{GitHub}.

The remainder of this paper is organized as follows. Section~\ref{sec:related} provides a comprehensive background in FL and its existing aggregation algorithms. In Section~\ref{sec:my_new_method}, we illustrate the detailed proposed method. In Section~\ref{sec:experiment}, we offer the experiment procedures as an evaluation process. Consequently, we demonstrate the experimental results in Section~\ref{sec:results}. Finally, we conclude this work and outline future research directions in Section~\ref{sec:conclusion}.
\section{Related Works}
\label{sec:related}
Our proposed approach addresses the critical limitations identified in traditional average and weighted mean aggregation methods in FL, particularly their failure to account for the heterogeneity of data quality.
Before we delve into the detailed discussion of our proposed method, it's necessary to offer a brief background of FL and, more importantly, the strengths and weaknesses of the existing aggregation methods.

FL represents a paradigm in collaborative machine learning, designed to address the growing concerns of privacy and data security in the digital age. Unlike traditional machine learning approaches that rely on centralizing data, FL enables the training of models across decentralized devices or servers while keeping the data localized, thus significantly mitigating privacy risks~\cite{Zheng2023, Gosselin2022} and exploiting edge computational resources.

The aggregation mechanism in FL is a pivotal concept that aggregates models across multiple decentralized nodes to generate a server-end global model. Existing aggregation approaches are not without limitations, particularly concerning the diversity and inconsistency of data across different clients, which can hinder the convergence and effectiveness of the global model. 
The following section elaborates further on the work related to the aggregation mechanism.

\subsection{Model Aggregation Mechanism}
Grudzie'n et al. highlights aggregation mechanisms' importance in reducing communication loads between servers and clients through compression and importance sampling, thereby enhancing FL's performance~\cite{Grudzien2023}. Similarly, Khan et al. underline the role of systematic collaboration selection and regularized weight aggregation in optimizing communication payloads, a testament to the cost-efficiency and practicality of advanced aggregation methods~\cite{Khan2023}.

Moreover, sophisticated aggregation mechanisms substantially enhance the resilience of FL systems against malicious attacks. Geng et al. introduce a novel approach, constructing a Byzantine-robust FL model that utilizes synthesized trust scores to enhance global model accuracy and defend against adversarial actions~\cite{Geng2023}. This is complemented by the work of Nguyen et al., who propose the FLAME framework to mitigate backdoor attacks through strategic noise injection, maintaining model integrity without compromising performance~\cite{Nguyen2021}.

The efficiency of FL also hinges on the aggregation mechanism's ability to reduce computation and communication costs. Islamov et al. discuss the integration of communication compression and Bernoulli aggregation in distributed Newton-type methods, showcasing a reduction in resource consumption while ensuring robustness~\cite{Islamov2022}. Furthermore, Yu et al. demonstrate the effectiveness of the G$^2$uardFL framework in identifying and neutralizing malicious clients through attributed client graph clustering, emphasizing the aggregation mechanism's role in enhancing security and model reliability~\cite{yu2023g}.

The decentralized infrastructure of FL, supported by private blockchain technology, presents a novel avenue for the secure and efficient transfer of clinical prediction models. Schapranow et al. illustrate this by detailing the NephroCAGE project, which employs FL for improved kidney transplantation outcomes, exemplifying the aggregation mechanism's capacity to protect sensitive data while facilitating local data analysis~\cite{Schapranow2023}.

\subsection{Popular Aggregation Algorithms}

In the landscape of FL, several aggregation methods have been proposed to integrate insights from distributed clients efficiently. The Simple Average Aggregation~\cite{McMahan2016CommunicationEfficientLO} treats each client's contribution equally, ignoring the size of their local datasets. This democratic approach, however, might expose the model to the risk of incorporating misleading data from less reliable sources. Conversely, the Weighted Mean Aggregation~\cite{reyes2021precision} allocates weights to client contributions based on the quantity of their data, ensuring that the aggregation reflects the proportional input of each client, thereby aiming for a more balanced aggregated outcome.

Another robust technique is the Median Aggregation~\cite{Pillutla2022}, which selects the median of the model parameters across all clients for aggregation. This method is particularly adept at mitigating the influence of outliers or erroneous data, enhancing the model's resilience to anomalies. Meanwhile, Momentum Aggregation~\cite{Xu2021FedCMFL} introduces a momentum term to smooth the aggregation over time by incorporating information from previous updates, which helps in accelerating convergence by acknowledging historical trends in parameter changes.

In a bid to personalize the learning process, Personalized Aggregation~\cite{9880164} merges the global model with an averaged update from all clients, aiming to better align the model with individual client peculiarities. Addressing privacy concerns, the Differential Privacy Average Aggregation~\cite{9069945} method integrates Laplace noise into the aggregation to protect participant data, balancing privacy with the attainment of a meaningful aggregated model. Lastly, the Quantization Aggregation (q-FedAvg)~\cite{9425020} seeks to diminish communication overhead by quantizing the model parameters before their aggregation, streamlining the update process while maintaining acceptable levels of accuracy.
\section{A Dual-Criterion Model Aggregation}\label{sec:my_new_method}

\subsection{Algorithm Overview}\label{subsec:improved_averaging_method}

This paper introduces a dual-weighting aggregation method that factors in both the quantity and quality of the data. The core contributions of our new method are twofold: i). We introduce a quality factor into the federated aggregation process, contrasting with traditional methods that rely solely on data quantity; ii). We employ a parameter $\lambda$ to dynamically adjust the weighting proportion between quality and quantity factors.

The basic working diagram is shown in Figure~\ref{fig:dual-criterion} with a more logical illustration in Algorithm~\ref{alg:main-flow}. Detailed information about the algorithm will be covered in the following section.

\begin{algorithm}[!h]
\SetAlgoLined 
\SetAlgoNlRelativeSize{-1} 
\SetNlSty{textbf}{}{} 
\newcommand{\MyCommentSty}[1]{\scriptsize\itshape {#1}}
\SetCommentSty{MyCommentSty} 
\caption{Dual-Criterion Model Aggregation}
\label{alg:main-flow}
Server initializes global model \(M(\theta^{0}_g)\) \label{line1};

\For{epoch \(k = 1\) to \(K\)}{
    \(M(\theta^{k-1}_g) \rightarrow M(\theta^{(i,k-1)}_l)\)\label{line3};
    
    \tcp{Server distributes the model to all clients}

    \vspace{0.5em}

    \textbf{Each client \(i\):} \\
    \Indp 
    Train local model \(M(\theta^{(i,k)}_l)\) on dataset \(D_i\);\label{line5}\\
    Compute \(Score_i\) and \(|D_i|\);\label{line6}\\
    Send \(\theta^{(i,k)}_l\), \(Score_i\), \(|D_i|\) to server;\label{line7}\\
    \Indm 

    \vspace{1em}
    
    \textbf{Server:} \\
    \Indp 
    Normalize \(|D_i|\) to \(v_i\);\label{line9}\\
    Normalize \(Score_i\) to \(q_i\);\label{line10}\\
    Compute weights \(f(q_i, v_i)\);\label{line11}\\
    \tcp{With dynamically adjustment of \(\lambda\)}
    Normalize weight \(f(q_i, v_i)\) to \(w_i\);\label{line12}\\

    \vspace{0.1em}

    Aggregate to generate global model \(M(\theta^{(k)}_g)\);\label{line13} \\
    \Indm 
}
\end{algorithm}

\subsection{Formula Analysis}
\label{subsec:formula}

In the framework of FL, the transmission of parameters is crucial; Hence, we represent the model's parameters as \(\theta\), then the global model at iteration \(k\) is denoted as \(M(\theta^k_g)\), and the local model trained by the \(i\)-th client at iteration \(k\) on data \(D_i\), is presented as \(M(\theta^{(i,k)}_l)\). Before each training round, the server first distributes the model to different clients as Line~\ref{line3} in Algorithm~\ref{alg:main-flow} shows.

In Line~\ref{line6}, after the local model training, we evaluate the local model to obtain the \(Score_i\), which represents the training quality of that client and can be regarded as the unnormalized version of the quality factor. The next challenge is to define \(Score_i\).


Given that the model structure across different clients is identical, it is feasible to evaluate the model performance on each client end, thereby indirectly reflecting the quality of the dataset of each client. In more straightforward terms, since the model structure is the same across different clients, the model trained on higher-quality datasets will perform better.

Based on this, we propose to assess the model's accuracy (notated as \(score_i\) in this paper), a most commonly used metric at the end of each epoch on every client, to serve as a reliable indicator of the model's generalization capability and, by extension, the quality of the training data used.



To implement this evaluation strategy, each client is required to keep an evaluation dataset that is not used for training but is instead used exclusively for model evaluation. The selection of this dataset should aim to cover a broad spectrum of the data distribution. In Line~\ref{line6}, we also simply record the size of each client dataset: \(|D_i|\).

Once \(Score_i\), \(|D_i|\), and the model parameters \(\theta^{(i,k)}_l\) are transmitted back to the server as Line~\ref{line7} shows, then in  the Line~\ref{line9}, quantity factor \(v_i\) is the normalized version of each client's dataset size \(|D_i|\):

\begin{equation}
v_i = \frac{|D_i|}{\sum_{j=1}^{N} |D_j|} 
\end{equation}

Same normalization for quality factor \(q_i\) is performed in Line~\ref{line10}:

\begin{equation}
q_i = \frac{Score_i}{\sum_{j=1}^{N} Score_j}
\label{subsec:q_iand_Score_i}
\end{equation}

Upon completion of the above procedures, we acquired the quality factor \(q_i\) and quantity factor \(v_i\). Now, we can compute the weights.

\begin{equation}
f(q_i, v_i) = \lambda \cdot q_i + (1 - \lambda) \cdot v_i
\label{equ:weight}
\end{equation}

\noindent where \(f(q_i, v_i)\) is the combination of \(q_i\) and \(v_i\), with dynamically adjustment of \(\lambda\). \(\lambda\) is a factor within \([0, 1]\) and is used to modulate the relative influence of quality over quantity in the final weight calculation. The choice of \(\lambda\) is not arbitrary; it is determined based on empirical evidence and the specific requirements of the FL task at hand. A higher value of \(\lambda\) privileges the quality of data, positing it as the primary determinant of a client's influence on the global model. Conversely, a lower value underscores the importance of data quantity, thereby amplifying the contributions from clients with larger datasets.

The rationale behind this weighted approach stems from the recognition that high-quality data can significantly enhance model accuracy, especially in scenarios where data is non-iid across clients. On the other hand, a larger quantity of data can also provide a more comprehensive representation of the problem space, facilitating model generalization.

For different application scenarios, you can define different values of \(\lambda\) to reflect your emphasis on data quality or quantity. However, in this paper, we have implemented a general approach that explores a range of values to optimize the model's accuracy and performance. In every machine learning algorithm, there is a validation set, after each epoch, the validation set is used to test the model, allowing the model with the best generalization performance to be stored. In our case, we use this validation set at the server end to assess the performance of models aggregated from different values of \(\lambda\), ultimately saving and using the best-performing \(\lambda\).

At the final stage in Lines~\ref{line12} and~\ref{line13}, as with the standard weighted aggregation procedure, the aggregated global model parameters \(\theta^k_g\) can be updated as follows:

\begin{equation}
\theta^k_g = \sum_{i=1}^{N} \left( w_i \cdot \theta^{(i,k)}_l \right)
\end{equation}

\noindent where \(w_i\) is the normalized weight:
\begin{equation}
w_i = \frac{f(q_i, v_i)}{\sum_{j=1}^{N} f(q_j, v_j)},
\end{equation}

\begin{table*}[t]
\captionsetup{font=footnotesize}
\caption{Aggregation Mechanisms in Our Experimental Setup}
\label{tab:Aggregation Mechanisms}
\resizebox{0.98\textwidth}{!}{ 
\begin{tabular}{m{0.25\textwidth}m{0.35\textwidth}m{0.40\textwidth}}  
\toprule
\textbf{Algorithm} & \textbf{Description} & \textbf{Formula} \\
\midrule
Simple Average Aggregation~\cite{McMahan2016CommunicationEfficientLO} & Averages all client model parameters equally. & 
$\theta^{\text{global}}_k = \frac{1}{N} \sum_{i=1}^{N} \theta^{(i)}_k$ \\ 
[1.2em]
Weighted Mean Aggregation~\cite{reyes2021precision} & Averages model parameters with weights proportional to the size of each client's local dataset. & 
$\theta^{\text{global}}_k = \sum_{i=1}^{N} w_i \cdot \theta^{(i)}_k$ \\ 
[1.2em]
Median Aggregation~\cite{Pillutla2022} & Takes the median value of each model parameter across all clients. & 
$\theta^{\text{global}}_k = \text{median}(\theta^{(1)}_k, \theta^{(2)}_k, \ldots, \theta^{(N)}_k)$ \\ 
[1.5em]
Momentum Aggregation~\cite{Xu2021FedCMFL} & Incorporates momentum into the global model updates to accelerate convergence. & 

$\mathbf{M}_k = \beta \mathbf{M}_{k-1} + \left( \sum_{i=1}^{N} \frac{n_i}{n} \theta^{(i)}_k - \theta^{\text{global}}_{k-1} \right) $ \newline

$\theta^{\text{global}}_k = \theta^{\text{global}}_{k-1} + \eta \mathbf{M}_k$\\ 
[2.5em]

Personalized Aggregation~\cite{9880164} & Takes a weighted average between the global model and the mean of all client updates. & 
$\theta^{\text{global}}_k = \alpha \theta^{\text{global}}_k + (1 - \alpha) \left( \frac{1}{N} \sum_{i=1}^{N} \theta^{(i)}_k \right)$ \\ 
[1.5em]
Differential Privacy Average~\cite{9069945} & It mainly adds Laplace noise to the aggregated model parameters. & 
$\theta^{\text{global}}_k = \left( \frac{1}{N} \sum_{i=1}^{N} \theta^{(i)}_k \right) + \mathcal{L}(0, \frac{1}{\epsilon})$ \\ 
[2.5em]

Quantization Aggregation~\cite{9425020} & Quantizes each client's model parameters before averaging. & 
$\theta^{(i, \text{quantized})}_k = \text{round}\left( \theta^{(i)}_k \times (2^{\text{Q Level}} - 1) \right) / (2^{\text{Q Level}} - 1)$ \newline
$\theta^{\text{global}}_k = \frac{1}{N} \sum_{i=1}^{N} \theta^{(i, \text{quantized})}_k$ \\
[2.5em]

\textbf{Our Proposed Aggregation} & Aggregates using weights that combine data quantity and quality, with adjustable parameter $\lambda$. & 
$v_i = \frac{|D_i|}{\sum_{j=1}^{N} |D_j|}, \quad q_i = \frac{\text{Score}_i}{\sum_{j=1}^{N} \text{Score}_j}$ \newline

$w_i = \lambda q_i + (1 - \lambda) v_i, \theta^{\text{global}}_k = \sum_{i=1}^{N} w_i \cdot \theta^{(i)}_k$ \\
[1.5em]
\bottomrule
\end{tabular}
}
\end{table*}

\subsection{Existing Aggregation Approaches}
To test the proposed aggregation algorithm, this work also delves into the examination of various established algorithms listed in Table~\ref{tab:Aggregation Mechanisms} which provides a more concise description of the existing aggregation mechanisms employed in our experimental setup. 

\section{Experimental Setting}
\label{sec:experiment}

After gaining an understanding of our algorithm from the preceding section, this section will describe our experimental procedure for the proposed approach. 

\subsection{Dataset}
\label{subsec:dataset}

As mentioned in the previous section, the proposed method requires a separate verification set to score each client model during training to assess the quality. In this paper, we define this specific verification set as the ``evaluation" dataset. And the term ``validation" dataset is consistently used to refer to the validation set which is used at the final stage to test and compare models.
Specifically, we evaluated our approach separately on two different datasets: the CIFAR-10 dataset and another synthetic dataset of robotic visual obstacle avoidance from our previous research efforts~\cite{yu2022federated}.

\subsubsection{CIFAR-10 dataset}
\label{subsubsec:Cifardataset}
The original CIFAR-10 dataset contains 50,000 training images and 10,000 testing images. The training set is divided into five training batches, with 10,000 images per batch. In our experimental setup, for the original 50000 training images, we assign the first four training batches to four separate clients, defined as C1-C4, the remaining 10,000 images are used as an evaluation set required for obtaining the quality factor \(q_i\). The original validation test set is retained for normal validation purposes.

\subsubsection{Visual Obstacle Avoidance Dataset}
\label{subsubsec:owndataset}

The distribution of the training datasets is delineated in Table~\ref{table:training_set}, where $C_i$ denotes the dataset associated with client $i$. Similarly, the evaluation set and validation set are collected as you can see in Table~\ref{table:validation_set}.

\begin{table}[h]
\centering
\captionsetup{font=footnotesize}
\caption{Training dataset distribution (visual obstacle avoidance datasets).}
\label{table:training_set}
\resizebox{0.9\linewidth}{!}{%
\begin{tabular}{lcccccc}
    \toprule
    Environment: & \multicolumn{2}{c}{Hospital ($C_1$)} & \multicolumn{2}{c}{Simple-room ($C_2$)} & \multicolumn{2}{c}{Warehouse ($C_3$)}\\
    Obstacle: & blocked & free & blocked & free & blocked & free \\
    \midrule
    Prop.& 46\% & 54\% & 48\% & 52\% & 59\% & 41\% \\
    \cmidrule(r){2-3} \cmidrule(lr){4-5} \cmidrule(l){6-7}
    Total & \multicolumn{2}{c}{272} & \multicolumn{2}{c}{217} & \multicolumn{2}{c}{397}\\
    \bottomrule
\end{tabular}
}
\end{table}

\begin{table}[h!]
\centering
\captionsetup{font=footnotesize}
\caption{Distribution of evaluation and validation data (visual obstacle avoidance datasets)}
\label{table:validation_set}
\resizebox{0.6\linewidth}{!}{%
\begin{tabular}{lcccc}
    \toprule
    & \multicolumn{2}{c}{Evaluation Dataset}  & \multicolumn{2}{c}{Validation Dataset}\\
    & blocked & free & blocked & free\\
    \midrule
    Prop.: & 47\% & 53\% & 47\% & 53\% \\
    \cmidrule(r){2-3} \cmidrule(lr){4-5}
    Total: & \multicolumn{2}{c}{452} & \multicolumn{2}{c}{510}\\
    \bottomrule
\end{tabular}
}
\end{table}

\subsection{Experimental Evaluation and Metrics}
\label{subsec:eva}

Throughout our experiments, we employed a range of metrics to evaluate the models, each offering unique perspectives on the model's capabilities and limitations. 

\begin{enumerate}
    \item \textbf{Accuracy} represents the ratio of correct predictions to the total number of predictions.
    \item \textbf{Precision}: quantifies the correctness of positive predictions, calculated as \(\frac{TP}{TP + FP}\). It is crucial in contexts where the consequences of false positives are severe, for example, it can ensure the reliability of the model in detecting obstacles.
    \item \textbf{F1 Score} is the harmonic mean of precision and recall. It balances the trade-off between precision and recall, providing a single metric that accounts for both false positives and false negatives.
    \item \textbf{Matthews Correlation Coefficient (MCC)} recognizes the limitations of accuracy and precision, especially in unbalanced datasets, the MCC is employed for its capacity to deliver a balanced assessment of the model's performance. Unlike other metrics, the MCC incorporates all four elements of the confusion matrix, offering a comprehensive evaluation by accounting for both the positive and negative classes.

\end{enumerate}

\subsection{Experiment Scheme}
\label{subsec:algorithm_framework}

\begin{algorithm}[!h]

\caption{Evaluation Scheme}
\label{alg:FL_Visual_Obstacle_Avoidance}
\footnotesize
\KwIn{\\
Datasets:\( D_{Client1}, D_{Client2}, D_{Client3},D_{Client4} \)\;
Client Combination List: \(\text{Combs}=[(\text{C1},\text{C2},\text{C3}), (\text{C1}, \text{C2},\text{C3}, \text{C4})]\)\;
Different Aggregation Methods List: \( \text{Agg\_methods} \)  \;
Training epochs: \( \text{Epochs} \)  \;
}

\KwOut{Trained federated models}

\BlankLine

\textbf{Initialization} \\

\( \text{Load train and validation datasets} \) \;
\( \text{Initialize client models and optimizers} \) \;
\BlankLine

\ForEach{\( \text{Comb} \in \text{Combs} \)}{
    \ForEach{\( \text{agg\_method} \in \text{Agg\_methods} \)}{
        \ForEach{\( \text{epoch} \in \text{range(Epochs)} \)}{
                \( \text{Distribute global model to client model} \) \;
            \ForEach{\( \text{client} \in \text{Comb} \)}{
                \( \text{Train the client model} \) \;
            }
            \( \text{Server-side model aggregation by agg\_method} \) \;
            \( \text{Test global model on validation data} \) \;
            \If{\( \text{performance improves} \)}{
                \( \text{Save global model} \) \;
            }
        }
    }
}

\textbf{Return trained federated models and performance metrics.}
\end{algorithm}


Our experiment setting, detailed in Algorithm~\ref{alg:FL_Visual_Obstacle_Avoidance}, is built around nested \texttt{foreach} loops. The outermost loop simulates different combinations of client datasets to see how various data sources affect the model's performance. Inside this, the next loop tests different aggregation methods---including our proposed method and other popular ones---to evaluate their impact on the results. The inner loops handle the core processes of the FL algorithm, such as synchronizing client models with the global model, running training passes, performing optimization steps, etc. It should be noted that the implementation details of different aggregation algorithms are not shown in this flowchart.

\subsection{Hardware and Software}
\label{subsec:hard_soft}
Experiments were run on a PC with an AMD R7 6800H processor and an Nvidia GeForce RTX 3050 GPU (4\,GB). The software environment consisted of Python 3.8, PyTorch 2.1.0, NumPy 1.24.3, sklearn 1.3.0, and Matplotlib 3.7.2, developed in Jupyter Notebook 6.5.4.
\section{Experimental Results}
\label{sec:results}

This section presents a comprehensive analysis of the experimental results obtained from various environmental combinations. As we mentioned above, the evaluation results are from the approaches in Table~\ref{tab:Aggregation Mechanisms} on two different datasets CIFAR-10 and visual obstacle avoidance specified dataset.


\begin{table*}[]
\centering
\captionsetup{font=footnotesize}
\caption{Results of CIFAR-10 Dataset from the aggregation of models from clients C1-C3 and C1-C4. The top-performing outcomes are indicated in bold, while our proposed method is underlined when it ranks as the second highest.}
\label{cifar-eval}
\begin{tabular}{l|cccc|cccc}
\toprule
\multirow{2}{*}{} & \multicolumn{4}{c|}{\textbf{C1\&C2\&C3}}                                & \multicolumn{4}{c}{\textbf{\textit{C1\&C2\&C3\&C4}}}     \\
& \multicolumn{1}{l}{Accuracy} & \multicolumn{1}{l}{Precision} & \multicolumn{1}{l}{F1 Score} & \multicolumn{1}{l|}{MCC} & \multicolumn{1}{l}{Accuracy} & \multicolumn{1}{l}{Precision} & \multicolumn{1}{l}{F1 Score} & \multicolumn{1}{l}{MCC} \\
\midrule
Dual-Criterion Weighted         & \textbf{0.8286} & \textbf{0.83239} & \textbf{0.82934} & \textbf{0.80983} & \textbf{0.8369} & \underline{0.8375}  & \textbf{0.83661} & \textbf{0.81891}  \\
Simple Average                  & 0.8199 & 0.8238  & 0.81962 & 0.80043 & 0.8301 & 0.83673 & 0.83107 & 0.81177  \\
Weighted Mean                   & 0.8226 & 0.82392 & 0.82084 & 0.80342 & 0.8303 & 0.83055 & 0.82981 & 0.81158  \\
Median                          & 0.8179 & 0.82109 & 0.81774 & 0.79805 & 0.8283 & 0.83008 & 0.82745 & 0.80962  \\
Momentum                        & 0.7772 & 0.7809  & 0.77613 & 0.75304 & 0.7797 & 0.78153 & 0.7773  & 0.75599  \\
Personalized                    &0.8257 & 0.82741 & 0.82581 & 0.80649 & 0.8342 & \textbf{0.83792} & 0.83419 & 0.81617  \\
Differential Privacy            & 0.1119 & 0.03877 & 0.05247 & 0.0159  & 0.1127 & 0.05479 & 0.05163 & 0.01855  \\
Quantization                    & 0.8227 & 0.82439 & 0.8221  & 0.80334 & 0.8291 & 0.83387 & 0.82832 & 0.81076  \\
\bottomrule
\end{tabular}
\end{table*}

\begin{table}[]
\centering
\captionsetup{font=footnotesize}
\caption{Results of Visual Obstacle Avoidance Dataset from the aggregation of models from clients C1-C2. The top-performing outcomes are indicated in bold, while our proposed method is underlined when it ranks as the second highest.}
\label{vsobstacle-eval}
\begin{tabular}{l|cccc}
\toprule
\multirow{2}{*}{} & \multicolumn{4}{c}{\textbf{C1\&C2\&C3}}        \\
& \multicolumn{1}{l}{Accuracy} & \multicolumn{1}{l}{Precision} & \multicolumn{1}{l}{F1 Score} & \multicolumn{1}{l}{MCC}  \\
\midrule
Dual-Criterion Weighted & \underline{0.99279} & \underline{0.99024} & \underline{0.99267} & \underline{0.98588} \\
Simple Average          & 0.98798 & 0.98068 & 0.98783 & 0.97606 \\
Weighted Mean           & 0.97596 & 0.95755 & 0.97596 & 0.95265 \\
Median                  & 0.93029 & 0.8821  & 0.93303 & 0.86711 \\
Momentum                & 0.96875 & 0.95694 & 0.96852 & 0.93778 \\
Personalized            & \textbf{0.99519} & \textbf{0.99029} & \textbf{0.99512} & \textbf{0.99043} \\
Differential Privacy    & 0.78365 & 0.86076 & 0.75138 & 0.57979 \\
Quantization            & 0.97837 & 0.96209 & 0.97831 & 0.95728 \\
\bottomrule
\end{tabular}
\end{table}

Before we delve into the detailed analysis of our methods, it is essential to briefly mention that in Table~\ref{cifar-eval} and Table~\ref{vsobstacle-eval}, the top-performing outcomes are indicated in bold, while our proposed method is underlined when it ranks as the second highest. 
In these results, our proposed method either achieves the best performance or demonstrates results comparable to the best, specifically due to the advantages of personalized training. Personalized training differs from the rest by incorporating additional local training after model aggregation, thereby enhancing the model's adaptability. 


For the CIFAR-10 dataset, as shown in Table~\ref{cifar-eval}, our dual-criterion weighted method consistently outperformed other aggregation methods. In the combination of clients C1, C2, and C3, our method achieved the highest accuracy of \(82.86\%\), F1 Score of \(82.93\%\), and MCC of \(80.98\%\). When all four clients (C1--C4) were included, it again led with an accuracy of \(83.69\%\), precision of \(83.75\%\), and MCC of \(81.89\%\). These results highlight our new method's effectiveness by integrating data quality and quantity into the aggregation process.
Compared to traditional methods like Simple Average and Weighted Mean, our approach showed notable improvements. For instance, in the C1--C3 setup, the Simple Average method yielded an accuracy of \(81.99\%\) and MCC of \(80.04\%\), both lower than our method's results.

In the visual obstacle avoidance dataset (Table~\ref{vsobstacle-eval}), our method demonstrated high efficacy, achieving an accuracy of \(99.28\%\), F1 Score of \(99.27\%\), and MCC of \(98.59\%\). These figures are comparable to the personalized aggregation, which suggests that our dual-criterion approach can effectively generalize to different types of data and tasks.

The superior performance of our method across both datasets underscores the advantage of considering data quality and quantity. By dynamically weighting client contributions based on these dual criteria, the aggregated global model benefits from more relevant and representative updates. This is particularly advantageous in FL scenarios characterized by data heterogeneity and non-iid distributions.

In summary, the experimental results validate our dual-criterion aggregation method as a robust approach that enhances model performance by effectively balancing the influence of data quality and quantity. This method holds significant promise for improving federated learning applications where data diversity and quality vary across clients.
\section{Conclusion}
\label{sec:conclusion}

This work introduced a new model aggregation algorithm that improves FL model performance by considering both data quantity and quality across client nodes. The experimental results confirm our algorithm's superiority, showing better performance and accuracy compared to conventional methods, on general-purpose datasets, CIFAR-10, and datasets specific for visual obstacle avoidance. It is worth noting that our approach outperforms or is comparable to the personalized training in the experiment.

Future work should focus on refining the algorithm. For example, we can explore the integration of advanced machine learning techniques, such as reinforcement learning or meta-learning, to dynamically adjust the weighting of data quality and quantity based on the learning context or performance metrics.


\section*{Acknowledgment}

This research is supported by the Research Council of Finland's Digital Waters (DIWA) flagship (Grant No. 359247) and AeroPolis project (Grant No. 348480), as well as the DIWA Doctoral Training Pilot project funded by the Ministry of Education and Culture (Finland). 

\bibliographystyle{unsrt}
\bibliography{bibliography}

\end{document}